\documentclass[letterpaper]{article} 
\usepackage{aaai2027}  
\usepackage[hyphens]{url}  
\usepackage{graphicx} 
\usepackage{natbib}  
\usepackage{caption} 
\usepackage{enumitem}
\usepackage{algorithm}
\usepackage{algorithmic}
\usepackage{amsthm} 
\newcommand{\std}[1]{\,\scriptsize{$\pm$ #1}}
\newtheorem{lemma}{Lemma}
\usepackage{pifont} 
\usepackage{newfloat}
\usepackage{amsfonts}
\usepackage{amsmath}
\usepackage{listings}
\DeclareCaptionStyle{ruled}{labelfont=normalfont,labelsep=colon,strut=off} 
\floatstyle{ruled}
\newfloat{listing}{tb}{lst}{}
\floatname{listing}{Listing}

\usepackage{booktabs}

\title{Distill Skills into Weights, Not Prompts: Abstract Skills as Privileged Signals for On-Policy Self-Distillation}
\author{
Yubo Jiang$^{1,2}$ \quad
\textbf{Fengying Xie}$^{1,3}$ \quad
\textbf{Zhiguo Jiang}$^{3}$ \quad
\textbf{Haopeng Zhang}$^{1,3~\dagger}$
}

\affiliations{
$^{1}$School of Astronautics, Beihang University, Beijing 102206, China\\
$^{2}$Longcat Interaction Team, Meituan, Beijing 100102, China\\
$^{3}$Tianmushan Laboratory, Beihang University, Hangzhou 311115, China\\[0.5em]
{\tt\small \{jbond0409, zhanghaopeng\}@buaa.edu.cn (Y.J., H.Z.)}
}

\begin{document}

\maketitle

\begin{abstract}
Reinforcement learning with verifiable rewards yields no
group-relative signal when rollout groups are uniformly correct or
uniformly wrong, which account for $63.0$--$68.0\%$ of groups in our
experiments. We propose SKALD (Skill-Anchored Latent Distillation), an
on-policy self-distillation framework that uses two context views of
the same Qwen3-Base model: a question-only student and a teacher
conditioned on an abstract, explicit-answer-filtered skill card. The
student is trained on its own prefixes, transferring the
skill-induced advantage into shared parameters without privileged
input at test time. To stabilize context-induced distribution
mismatch, SKALD employs an annealed exponentially tilted objective
that downweights teacher-preferred tokens with very low student
likelihood; as the tilt vanishes, it converges to teacher
cross-entropy and recovers the forward-KL student gradient. An
empirical gate activates distillation only when verified rollouts
estimate a positive teacher advantage. Across five held-out
mathematics benchmarks, SKALD improves overall avg@8 over GRPO by
$+2.46$, $+4.85$, and $+12.01$ at 0.6B, 1.7B, and 4B, respectively.
At 1.7B, zero-variance-only distillation recovers $84.7\%$ of the full
gain, while SKALD remains $+4.06$ above FLOP-matched GRPO and exceeds
contextual skill exposure by $+3.77$. These results show that
abstract skills provide dense supervision where group-relative
rewards become uninformative.
\end{abstract}
\section{Introduction}
Reinforcement learning with verifiable rewards (RLVR) has become central to improving mathematical reasoning in large language models, yet its outcome supervision is sparse \cite{chen2026learning,shao2024deepseekmath,wu2025invisible,yu2026dapo}. In group-relative policy optimization, identical rewards within a rollout group produce zero group-normalized advantages, so that group contributes no reward-gradient signal on the corresponding problem. This algebraic blind spot can occur for both all-correct and all-wrong groups, including difficult problems on which sampled solutions are uniformly unsuccessful~\cite{sun2024logit,xu2024survey}.

On-policy distillation is a natural complement: the student samples its own rollouts while a teacher distribution supplies dense token-level supervision along the sampled trajectory even when outcome rewards do not discriminate. The teacher need not be a separate online model~\cite{wu2025invisible,ye2026policy}: privileged context can make another view of the same parameters more informative on a given problem, and distillation can transfer that context-induced shift into the deployable question-only branch~\cite{li2026rethinking,fu2026revisiting,yang2026learning,zhao2026selfdistilledreasoneronpolicyselfdistillation}.

\begin{figure*}[t]
    \centering
    \includegraphics[width=0.8\textwidth]{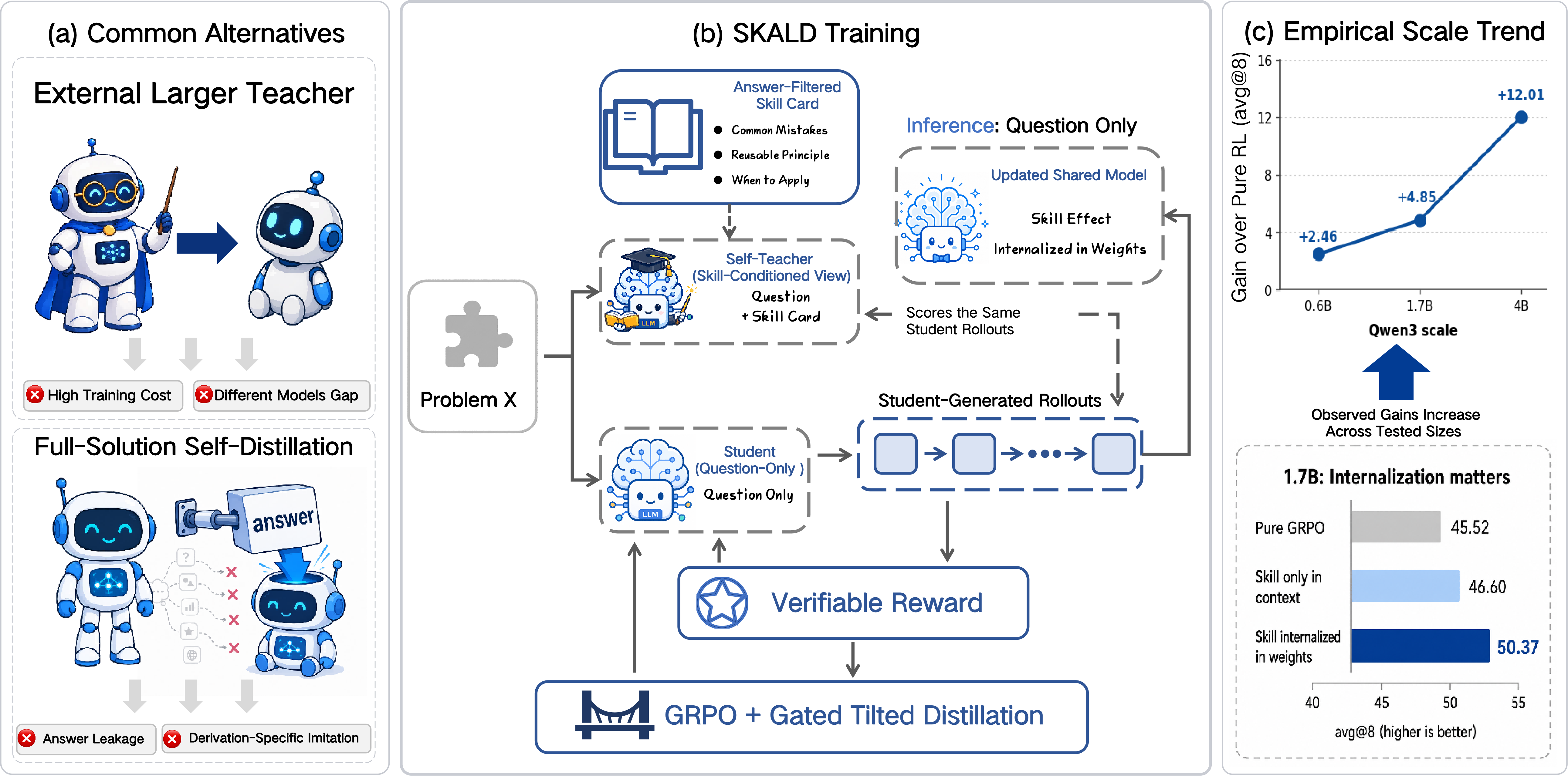}
\caption{
(a) Common privileged-distillation designs use a larger online teacher or expose a full reference solution to the teacher branch.
(b) SKALD instead transfers an explicit-answer-filtered skill through shared-parameter self-distillation.
(c) Across the three tested Qwen3-Base sizes, the observed gain over GRPO increases from $+2.46$ at 0.6B to $+12.01$ at 4B; this is an empirical trend, not a claimed scaling law.
}
\label{fig:skald_overview}
\end{figure*}
The central question is what privileged information to provide. A full reference solution exposes the final answer and a particular derivation to the teacher branch; a larger online teacher avoids that input but adds training and deployment dependencies. SKALD instead conditions the self-teacher on a compact skill abstraction---a relevant principle, when it applies, and common mistakes to avoid---while the student sees the question alone. Both branches score the same student-generated prefixes. At test time, neither the skill nor a teacher branch is required. The method does use a larger model offline to construct and screen cards; our ``no larger teacher'' claim therefore refers only to online distillation and inference.This design contrasts skill \emph{use} with skill \emph{internalization}: a skill in the student prompt acts as a temporary scaffold, whereas teacher-only distillation changes the parameters used by the question-only policy~\cite{shi2022skill,nam2022skill,xu2026agent}. In the 1.7B no-skill-at-test comparison, GRPO reaches $45.52$, training-time contextual exposure reaches $46.60$, and SKALD reaches $50.37$. Thus SKALD is $+3.77$ above contextual exposure in this setting; the comparison supports a weight-level transfer effect but does not by itself isolate every training difference.

Direct teacher cross-entropy can be unstable when a privileged context assigns substantial mass to tokens the student currently considers improbable~\cite{wang2026skill,he2026self}. SKALD uses the annealed objective $-\tau^{-1}\log\mathbb{E}_{v\sim q}[p(v)^{\tau}]$. At $\tau>0$, it creates a student-dependent effective target that discounts very low-$p$ teacher tokens and has the sharpened stationary solution $p^*\propto q^{1/(1-\tau)}$; annealing $\tau:0.8{\to}0$ removes this transient mode-seeking bias and recovers teacher cross-entropy, equivalently the forward-KL student gradient under stop-gradient. A fixed empirical gate further filters distillation using an initial sampled accuracy difference; it is a noisy heuristic, not a certificate of teacher superiority throughout training.
Across Qwen3-Base models at 0.6B, 1.7B, and 4B, SKALD outperforms GRPO on the problem-weighted average of MATH500~\cite{lightman2024let}, AMC23~\cite{amc23}, AIME24~\cite{aime24}, AIME25~\cite{aime25}, and Minerva~\cite{lewkowycz2022solvingquantitativereasoningproblems}, with observed gains of $+2.46$, $+4.85$, and $+12.01$. Three sizes from one model family establish an encouraging monotonic trend, not a general scaling law. Our contributions are:
\begin{itemize}[itemsep=2pt, topsep=3pt, leftmargin=*]
    \item We introduce shared-parameter, on-policy skill distillation that transfers an explicit-answer-filtered abstraction into a question-only policy without a larger online teacher or test-time privileged input.
    \item We characterize the tilted objective as a R\'enyi-type cross-entropy with a student-dependent escort target, bounded logit-gradient coordinates, and a cross-entropy limit whose student gradient matches forward KL; the contribution is its annealed use under privileged on-policy mismatch.
    \item We directly localize the gain to zero-variance rollout groups and validate it with conditional-leakage audits, problem/solution/skill decontamination, hierarchical uncertainty, equal-compute GRPO, gate sensitivity, and a full 4B component ablation.
\end{itemize}
\begin{figure*}[t]
    \centering
    \includegraphics[width=0.8\textwidth]{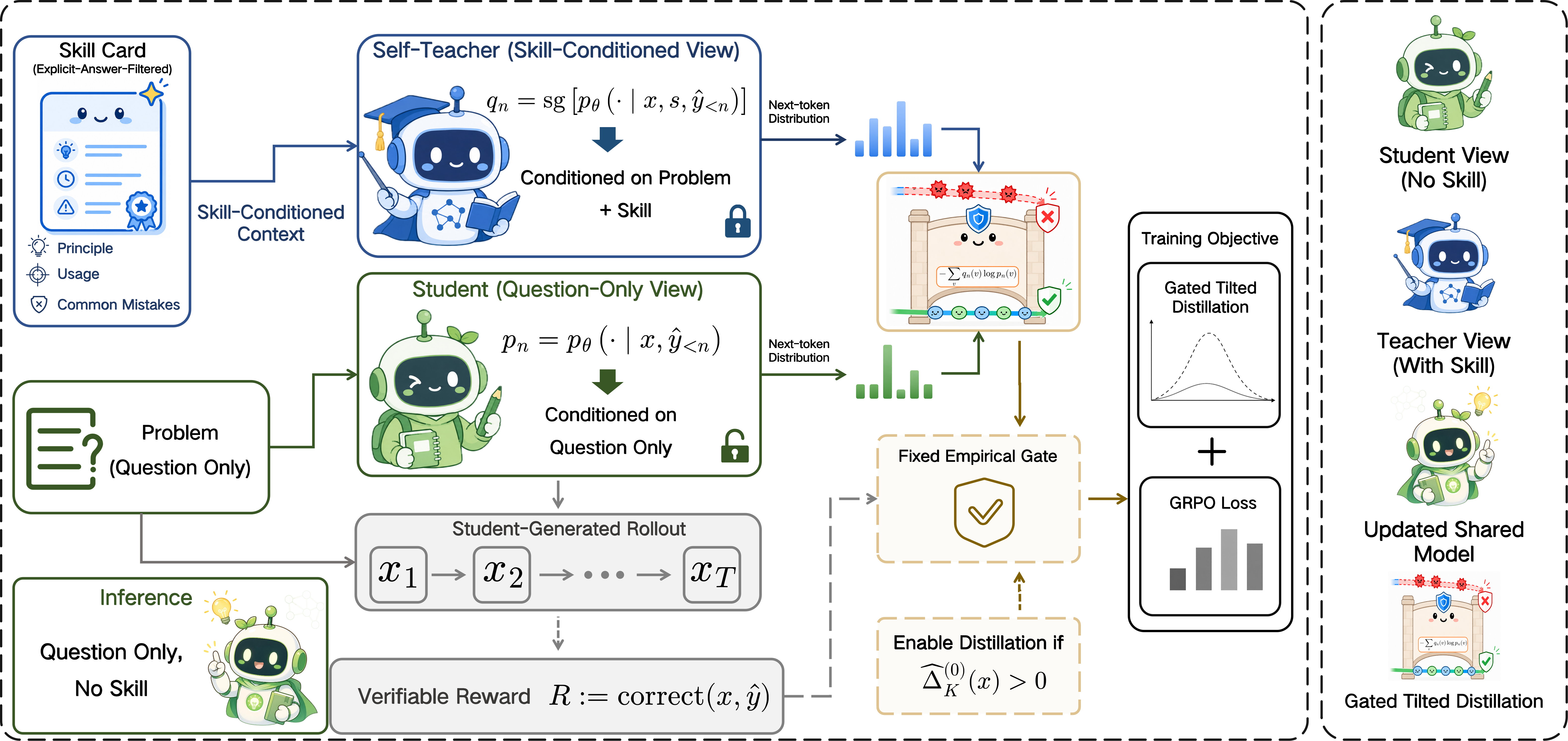}
\caption{
SKALD overview. The skill-conditioned teacher and question-only
student branches are two context views of the same Qwen3-Base
checkpoint and share all parameters. The teacher distribution is
evaluated under stop-gradient, while gradients flow through the
student branch. For $\tau>0$, the tilted objective induces the
effective target $r_\tau(v)\propto q(v)p(v)^\tau$, suppressing
teacher-preferred tokens with very low student likelihood; it
therefore does not preserve $q$ as the stationary solution. A fixed
empirical gate activates distillation when initial sampled teacher
accuracy exceeds student accuracy. At inference, the model receives
only the question.
}
\label{fig:skald_method}
\end{figure*}
\section{Related Work}
SKALD sits at the intersection of four threads: on-policy distillation, learning with privileged information, stabilization under distribution mismatch, and skill-based reasoning.
\paragraph{On-policy distillation.}
Classical sequence-level distillation trains on teacher corpora, inducing a train--inference mismatch~\cite{kim2016sequence,agarwal2024policy,ye2026policy}. GKD removes this mismatch by distilling on the student's own samples and composes naturally with RL~\cite{ye2022generalized,zhao2026selfdistilledreasoneronpolicyselfdistillation}; at $\tau=0$, SKALD reduces to on-policy teacher cross-entropy (and hence the forward-KL student gradient); its primary departure is the \emph{source} of supervision. Many GKD-style methods use an external larger teacher, creating a capacity gap and additional training cost~\cite{gou2021knowledge,cho2019efficacy}. SKALD's online teacher is the same weights in a privileged context and adds no inference-time dependency~\cite{tian2025knowledge,yang2025survey,mansourian2025comprehensive}, but it still requires an extra teacher-scoring forward pass during training; end-to-end accounting is in Supplementary Section~2.6.
\paragraph{Privileged self-teachers.}
Learning using privileged information lets a teacher observe training-only features the student never will~\cite{vapnik2009new,lopez2015unifying}. Its instantiation for reasoning LLMs is OPSD~\cite{zhao2026selfdistilledreasoneronpolicyselfdistillation}, whose teacher branch sees the per-instance reference solution, final answer included~\cite{zhao2026self}. Full solutions expose the answer and one reference derivation to the teacher branch and can increase teacher--student mismatch~\cite{penaloza2026privileged}. SKALD supplies an explicit-answer-filtered skill instead---a principle, its applicability, and common mistakes---and contains the OPSD objective as a special case: setting $s=y^{*}$ and $\tau=0$ recovers its objective, with skill abstraction and risk tilting independently ablatable~\cite{garcia2019learning}. $\pi$-Distill~\cite{penaloza2026privileged} trains a privileged-conditioned teacher and an unconditioned student in shared parameters as parameter-shared variational EM; SKALD instead makes the privileged signal an explicit, discrete, library-backed skill, estimates its usefulness per problem with an outcome-based
empirical gate, and internalizes it via annealed on-policy cross-entropy rather than implicit variational inference~\cite{liu2025efficient}.
\paragraph{Generalized cross-entropies and mismatch.}
Equation~\eqref{eq:tilt} is not a new divergence in isolation. With
order $\alpha=1+\tau$, it is a R\'enyi cross-entropy of the teacher
with respect to the student under one established definition, and
can equivalently be viewed as the negative logarithm of a
teacher-weighted generalized power mean~\cite{valverde2019shiftingrenyi,thierrin2022renyicrossentropy}. It is related to, but distinct from,
standard R\'enyi/Chernoff divergences, whose power exponents are
coupled to sum to one~\cite{vanerven2014renyi}. It also differs from
standard temperature distillation, which explicitly rescales logits:
here the teacher probabilities are not rewritten, but the optimization
induces the student-dependent effective target
$r_\tau(v)\propto q(v)p(v)^{\tau}$ and the stationary escort
$p^*\propto q^{1/(1-\tau)}$. We therefore claim novelty for the
shared-context, privileged on-policy construction and its annealed use,
not for the power-mean functional itself.
\paragraph{Skills: context versus weights.}
A parallel line treats natural-language skills as first-class objects: SkillRL~\cite{xia2026skillrlevolvingagentsrecursive} evolves a hierarchical library of reusable skills alongside the policy, and SKILL0~\cite{lu2026skill0incontextagenticreinforcement} places skills in the policy context during training and withdraws them under a helpfulness curriculum~\cite{snell2022learningdistillingcontext,laskin2022incontextreinforcementlearningalgorithm,ye2026onpolicycontextdistillationlanguage}. Both keep the skill on the student's input, so the model reasons \emph{with} the scaffold and must survive its removal. SKALD takes the complementary route: the skill appears only in the teacher, never in the student's context, and distillation writes its effect directly into the weights. Under a no-skill-at-test protocol the two routes are comparable, and our no-skill-at-test experiment shows that weight-level transfer is $+3.77$ points above contextual exposure with withdrawal (\S4.4).
\section{Method}
SKALD trains a single model through two context views. Skill conditioning can, but need not, improve the model's conditional distribution on a given problem; the empirical gate is designed to filter cases in which a sampled improvement is not observed. A skill-conditioned branch scores the rollouts of a question-only branch, and an annealed cross-entropy objective transfers their distributional difference into the shared weights. Figure~\ref{fig:skald_method} gives an overview. We present the two-context setup (\S3.1), the tilted cross-entropy and its effective target (\S3.2), the full objective (\S3.3), and the method's design desiderata and scope (\S3.4).
\subsection{Setup: One Model, Two Contexts}
For each scale, a single Qwen3-Base checkpoint $p_\theta$ is
instantiated as two context views that share all parameters. The student $p_S(\cdot\mid x):=p_\theta(\cdot\mid \mathrm{render}_S(x))$ sees only the problem $x$ and matches the deployment condition exactly; the self-teacher $p_T(\cdot\mid x,s):=p_\theta(\cdot\mid \mathrm{render}_T(x,s))$ additionally sees an abstract skill $s$. This construction eliminates the two standing costs of external teachers at once: there is no capacity mismatch, because teacher and student share every parameter, and no serving dependency, because the teacher exists only as an alternative prompt~\cite{ye2026policy,zhang2026opsdl}.
All trajectories $\hat{y}$ are sampled from the student. For each
position, define
$p_n:=p_S(\cdot\mid x,\hat{y}_{<n})$ and
$q_n:=\operatorname{sg}[p_T(\cdot\mid x,s,\hat{y}_{<n})]$,
where $\operatorname{sg}$ denotes stop-gradient. This operation freezes the teacher only within one update: because the branches share parameters, the teacher distribution changes after every optimizer step. SKALD is therefore a coupled moving-target self-distillation procedure, not distillation from a fixed teacher. During optimization, the teacher scores student rollouts in one prefill pass; separate teacher rollouts are used only to precompute the usefulness gate. This asymmetry keeps all distillation prefixes
on-policy under the student, avoiding supervision on
teacher-generated prefix distributions.
The skill is a structured card
$s=\{\texttt{title},\texttt{principle},\texttt{usage},\texttt{common mistakes}\}$
mined offline from training solutions. The pipeline retains $11{,}382$ cards
from $17{,}384$ candidates and merges them into $1{,}926$ reusable skills;
a skill is linked to a median of $5$ problems (mean $7.4$), and $93.1\%$ of
training problems receive an assignment. A larger Qwen3-14B model is used
only for offline extraction and screening, not online distillation or
inference. In a conditional audit, answer recovery is $2.4\%$ from the
question alone, $3.7\%$ with a shuffled skill, $5.1\%$ with the matched
skill, and $94.2\%$ with the full solution; removing instance-specific
numerals, entities, and intermediate expressions changes avg@8 only from
$50.37$ to $50.24$. These controls substantially narrow, but cannot
logically eliminate, conditional-leakage concerns (Supplementary
Section~1.3).

Boxed-answer extraction yields the binary reward
$R:=\operatorname{correct}(x,y)$. The default gate is precomputed from the
initial shared checkpoint using $K=8$ rollouts per branch and remains fixed.
It is an empirical filter, not a certificate: it covers $61.2\%$ of training
problems, and $14.6\%$ of initially positive cases reverse sign by training's
end. Nevertheless, hard/soft and fixed/refreshed variants score within
$49.98$--$50.48$ at 1.7B (Supplementary Section~2.4)~\cite{liao2025skintern,jiang2026sok}.
\subsection{Annealed Tilted Cross-Entropy}
Direct teacher cross-entropy can be dominated by positions at which the
teacher assigns mass to tokens that the student currently considers highly
improbable. Let $\ell_n(v):=\log p_n(v)$ and
$d_n(v):=-\ell_n(v)$. For $\tau>0$, SKALD uses
\begin{equation}
\begin{aligned}
\mathcal{L}^{(\tau)}_n
&=-\frac{1}{\tau}\log
\mathbb{E}_{v\sim q_n}\!\left[e^{\tau\ell_n(v)}\right]\\
&=-\frac{1}{\tau}\log\sum_v q_n(v)p_n(v)^{\tau}.
\end{aligned}
\label{eq:tilt}
\end{equation}
We define $\mathcal{L}^{(0)}_n:=H(q_n,p_n)$ by continuity. Thus
\begin{equation}
\lim_{\tau\to0}\mathcal{L}^{(\tau)}_n
=H(q_n,p_n)
=\mathrm{KL}(q_n\Vert p_n)+H(q_n).
\label{eq:limit}
\end{equation}
Because $q_n$ is stop-gradient within an update, the limit has exactly the
same student gradient as forward KL, but its numerical value differs by the
teacher entropy $H(q_n)$. For small positive $\tau$, the cumulant expansion
is
\begin{equation}
\mathcal{L}^{(\tau)}_n
=\mathbb{E}_{q_n}[d_n]
-\frac{\tau}{2}\operatorname{Var}_{q_n}[d_n]
+O(\tau^2).
\label{eq:cumulant}
\end{equation}
The negative variance correction is not a variance penalty. It is an
optimistic entropic risk for the loss $d_n$: high-disutility tokens are
discounted relative to ordinary cross-entropy.

\begin{lemma}[Effective Target, Logit Gradient, and Stationarity]
\label{lem:tilt}
Let
\[
r_n^{(\tau)}(v)
:=\frac{q_n(v)p_n(v)^{\tau}}
{\sum_u q_n(u)p_n(u)^{\tau}}.
\]
If $z_n$ denotes the student logits, then
\[
\nabla_{z_n}\mathcal{L}^{(\tau)}_n
=p_n-r_n^{(\tau)}.
\]
Consequently, each logit-gradient coordinate lies in $[-1,1]$.
A bound on $\|\nabla_\theta\mathcal{L}^{(\tau)}_n\|$ additionally
requires a bound on the logit Jacobian with respect to $\theta$.
For $0<\tau<1$, the minimizer over the probability simplex is
\[
p_n^*(v)=
\frac{q_n(v)^{1/(1-\tau)}}
{\sum_u q_n(u)^{1/(1-\tau)}}.
\]
At $\tau=0$, $r_n^{(0)}=q_n$ and $p_n^*=q_n$.
\end{lemma}
A complete proof is provided in Supplementary Section~3.

Lemma~\ref{lem:tilt} makes the optimization bias explicit. At the initial
$\tau=0.8$, the stationary distribution is proportional to $q_n^5$, a
strongly sharpened escort rather than the original teacher distribution.
We use this mode-seeking bias only transiently: as the student's support
improves and $\tau$ is annealed to zero, the effective target approaches
$q_n$ and the objective approaches teacher cross-entropy. The extra arithmetic is small once teacher and student logits are
available, but SKALD as a whole is not free relative to GRPO. We use
$\lambda_R=\lambda_D=1$ at all scales; $\tau=0.8$ for the first $10\%$ of
$370$ updates, decay it linearly to $0$ by $80\%$, and keep $\tau=0$ for the
last $20\%$. Learning rates are $2\!\times\!10^{-6}$, $1\!\times\!10^{-6}$,
and $5\!\times\!10^{-7}$ at 0.6B, 1.7B, and 4B, respectively
(Supplementary Sections~1.1 and~2.5).

\subsection{Full Objective and Reductions}
The training loss combines the verifiable reward with gated skill distillation:
\begin{equation}
\mathcal{L}(\theta)=\lambda_R\,\mathcal{L}_R(\theta)
+\lambda_D\,\mathbb{E}_x\!\Big[g_K(x)\,\mathbb{E}_{\hat{y}\sim p_S}
\tfrac{1}{|\hat{y}|}\!\sum_n
\mathcal{L}^{(\tau)}_n\Big],
\label{eq:loss}
\end{equation}
where $\mathcal{L}_R$ is group-normalized REINFORCE
(GRPO)~\cite{shao2024deepseekmath} on student rollouts and
\begin{equation}
\begin{aligned}
g_K(x)
&=
\mathbf{1}\!\left[
\widehat{\Delta}^{(0)}_K(x)>0
\right],
\\
\widehat{\Delta}^{(0)}_K(x)
&=
\frac{1}{K}\sum_{i=1}^{K}
\left[
R(y_i^T)-R(y_i^S)
\right].
\end{aligned}
\end{equation}
with
$y_i^T\sim p_T^{(0)}(\cdot\mid x,s)$ and
$y_i^S\sim p_S^{(0)}(\cdot\mid x)$.
The superscript $(0)$ denotes the initial shared checkpoint; the
resulting gate remains fixed during training. It selects problems for which
the sampled initial difference is positive, but finite-$K$ error and later
teacher drift can create false positives, false negatives, and sign
reversals. On a rollout group with identical rewards, the GRPO term has zero
group-relative gradient, whereas distillation can remain nonzero whenever
$g_K(x)=1$ and $r_n^{(\tau)}\neq p_n$. The group-stratified experiment in
\S4.2 directly tests whether this algebraic complementarity explains the
observed gain.
We use the forward direction to retain teacher support, but at $\tau>0$
the effective target is sharpened and is not ordinary forward KL.
Setting $s=y^*$, $\tau=0$, $g\equiv1$, and $\lambda_R=0$ yields the OPSD
student gradient because $H(q,p)$ and $\mathrm{KL}(q\Vert p)$ differ only
by the stop-gradient teacher entropy. Dropping the distillation term yields
GRPO. The ordered ablation in \S4 studies skill abstraction, tilting, and
the gate; its marginal increments can depend on the order and on component
interactions.
\subsection{Design Desiderata and Scope}
Skill distillation is most plausible under three measurable desiderata.
First, \emph{reliability}: skill conditioning should improve expected
rollout accuracy,
\[
\Delta^*(x)=
\mathbb{E}_{y^T\sim p_T}[R(y^T)]-
\mathbb{E}_{y^S\sim p_S}[R(y^S)]>0.
\]
The fixed gate estimates this quantity only at initialization and with
finite samples. Second, \emph{nontriviality}: the skill should change the
next-token distribution on reasoning-relevant positions rather than only
boilerplate. Third, \emph{manageable mismatch}: the privileged context
should avoid an extreme teacher--student support gap. These are diagnostic
desiderata, not sufficient conditions for policy improvement, and the
shared-parameter teacher may cease to satisfy them as training evolves.
Figure~\ref{fig:h1} reports token-level and mismatch diagnostics;
Supplementary Sections~2.4 and~2.7 report gate refresh and group-stratified
controls. Separate frozen/EMA-teacher trajectories remain an open diagnostic
for the shared moving-target dynamics.

The three tested Qwen3-Base sizes show a monotonic increase in SKALD's
advantage over GRPO. This observation is consistent with the hypothesis
that more capable models can exploit the same skill card more effectively,
but three points from one model family do not establish a scaling law or a
model-family-independent property.
\begin{table*}[t]
\centering
\small
\setlength{\tabcolsep}{5.5pt}
\renewcommand{\arraystretch}{0.8}
\begin{tabular}{lcccccc}
\toprule
 & MATH500 & AMC23 & AIME24 & AIME25 & Minerva & Overall \\
Method & \multicolumn{6}{c}{avg@8\;/\;pass@8} \\
\midrule
\multicolumn{7}{c}{\textit{Qwen3-4B-base~\cite{yang2025qwen3technicalreport}}} \\
\midrule
Zero-shot
& 69.70\,/\,88.40
& 48.12\,/\,82.50
& 9.58\,/\,20.00
& 6.67\,/\,23.33
& 25.55\,/\,47.06
& 50.70\,/\,70.64 \\
Few-shot ($k{=}2$)
& 56.97\,/\,88.40
& 37.50\,/\,77.50
& 10.00\,/\,20.00
& 7.92\,/\,20.00
& 25.05\,/\,48.90
& 42.82\,/\,70.87 \\
SFT (GT-CoT)
& 64.30\,/\,86.20
& 38.44\,/\,77.50
& 5.00\,/\,13.33
& 4.17\,/\,20.00
& 31.20\,/\,48.90
& 48.68\,/\,69.38 \\
GRPO
& 61.85\,/\,83.00
& \textbf{57.81}\,/\,80.00
& 14.58\,/\,33.33
& 15.00\,/\,26.67
& 39.25\,/\,54.41
& 51.37\,/\,70.30 \\
SKALD (ours)
& \textbf{82.33}\,/\,94.00
& 52.81\,/\,85.00
& \textbf{20.00}\,/\,46.67
& \textbf{18.33}\,/\,36.67
& \textbf{39.84}\,/\,55.51
& \textbf{63.38}\,/\,77.98 \\
\midrule
\multicolumn{7}{c}{\textit{Qwen3-1.7B-base~\cite{yang2025qwen3technicalreport}}} \\
\midrule
Zero-shot
& 54.40\,/\,83.60
& 32.20\,/\,70.00
& 2.92\,/\,13.33
& 3.33\,/\,6.67
& 21.00\,/\,43.00
& 39.43\,/\,65.25 \\
Few-shot ($k{=}2$)
& 46.30\,/\,73.70
& 25.10\,/\,65.00
& 2.50\,/\,3.33
& 0.00\,/\,0.00
& 11.80\,/\,26.80
& 31.47\,/\,53.72 \\
SFT (GT-CoT)
& 51.30\,/\,76.20
& 23.40\,/\,62.50
& 2.50\,/\,6.67
& 0.00\,/\,0.00
& 22.80\,/\,43.10
& 37.69\,/\,60.23 \\
GRPO
& 62.50\,/\,83.20
& 35.90\,/\,67.50
& 4.58\,/\,16.67
& 4.17\,/\,16.67
& 24.80\,/\,43.00
& 45.52\,/\,65.36 \\
GRPO + skill in-ctx (withdrawn)
& 63.40\,/\,83.40
& 38.80\,/\,70.00
& 2.92\,/\,13.33
& \textbf{6.67}\,/\,23.33
& 26.10\,/\,43.40
& 46.60\,/\,65.83 \\
OPSD~\cite{zhao2026selfdistilledreasoneronpolicyselfdistillation} (full solution)
& 66.80\,/\,85.10
& 35.90\,/\,50.00
& 5.42\,/\,16.67
& 5.00\,/\,16.67
& 26.30\,/\,41.20
& 48.51\,/\,65.09 \\
EOPD~\cite{jin2026entropyawareonpolicydistillationlanguage} (entropy routing)
& 64.10\,/\,84.80
& 39.20\,/\,67.50
& 7.08\,/\,20.00
& 4.17\,/\,20.00
& 26.90\,/\,43.10
& 47.33\,/\,66.54 \\
SKALD (ours)
& \textbf{68.80}\,/\,86.40
& \textbf{42.20}\,/\,75.00
& \textbf{9.58}\,/\,26.67
& 5.00\,/\,20.00
& \textbf{27.20}\,/\,44.90
& \textbf{50.37}\,/\,68.59 \\
\midrule
\multicolumn{7}{c}{\textit{Qwen3-0.6B-base~\cite{yang2025qwen3technicalreport}}} \\
\midrule
Zero-shot
& 35.20\,/\,68.60
& 15.31\,/\,50.00
& 1.67\,/\,3.33
& 0.83\,/\,3.33
& 7.08\,/\,23.53
& 23.19\,/\,49.20 \\
Few-shot ($k{=}2$)
& 16.00\,/\,48.60
& 4.69\,/\,25.00
& 0.00\,/\,0.00
& 0.00\,/\,0.00
& 5.42\,/\,18.38
& 11.08\,/\,34.75 \\
SFT (GT-CoT)
& 36.73\,/\,66.60
& 16.88\,/\,47.50
& 1.67\,/\,3.33
& 0.42\,/\,3.33
& \textbf{11.67}\,/\,29.78
& 25.55\,/\,49.89 \\
GRPO
& 44.32\,/\,69.60
& 19.06\,/\,57.50
& 1.67\,/\,3.33
& 0.83\,/\,3.33
& 8.04\,/\,23.90
& 28.88\,/\,50.23 \\
SKALD (ours)
& \textbf{46.75}\,/\,71.20
& \textbf{19.38}\,/\,70.00
& 1.67\,/\,3.33
& \textbf{2.50}\,/\,10.00
& 11.21\,/\,26.10
& \textbf{31.34}\,/\,52.64 \\
\bottomrule
\end{tabular}
\caption{Per-benchmark avg@8 / pass@8 across scales. Overall is the problem-count-weighted micro-average over 872 problems, computed from unrounded per-problem results. Reported numbers are means over 3 independent training seeds; per-seed values and cross-seed deviations are in Supplementary Sections~2.1 and~2.3. Hierarchical intervals jointly resampling seeds and problems are reported in Supplementary Section~2.1.}
\label{tab:main}
\end{table*}
\section{Experiments}
\subsection{Setup}
We evaluate MATH500~\cite{lightman2024let} (500 problems),
AMC23~\cite{amc23} (40), AIME24~\cite{aime24} (30),
AIME25~\cite{aime25} (30), and Minerva~\cite{lewkowycz2022solvingquantitativereasoningproblems}
(272), totaling $872$ problems. Each checkpoint uses eight samples at
$T=0.7$, top-$p=0.95$; avg@8 is primary and overall scores are
problem-weighted micro-averages. Training uses DAPO-Math-17k and the
mathematics split of OpenThoughts. Exact, formula-normalized, and semantic
checks flag unions of $3$ problem texts, $14$ reference solutions, and $10$
skill cards against the evaluation suite; every reported model is retrained
after removing the union, leaving $23{,}691$ problems (Supplementary
Section~1.4).

The conditional leakage audit uses a common judge and decoding protocol:
question-only, empty-card, random-text, shuffled-skill, matched-skill, and
full-solution recovery rates are $2.4$, $2.5$, $2.6$, $3.7$, $5.1$, and
$94.2\%$, respectively. A three-rater audit finds no full-answer leak in
$200$ retained cards ($\kappa=0.83$), and the sanitized-card score is
$50.24$ versus $50.37$ for the original. Median/95th-percentile normalized
solution--card overlap is $0.03/0.11$ (Supplementary Section~1.3).

At 1.7B, SKALD-shuffled reaches $48.83$: $+3.31$ over GRPO, $+0.32$ over
OPSD, and $-1.54$ below matched SKALD. Thus matching matters, but generic
card exposure or regularization can explain part of the gain. We report
three training seeds and use a two-level bootstrap that first resamples
seeds and then evaluation problems within seed; decoding samples are
aggregated at the problem level. Compute accounting includes scoring and
gate construction, and equal-FLOP controls are reported in Supplementary
Sections~2.1 and~2.6. Full implementation, baseline, and per-benchmark
statistics are in Supplementary Sections~1, 2.2, and~2.3.

\subsection{Main Results and Mechanism Test}
\begin{figure}[t]
\centering
\includegraphics[width=0.98\columnwidth]{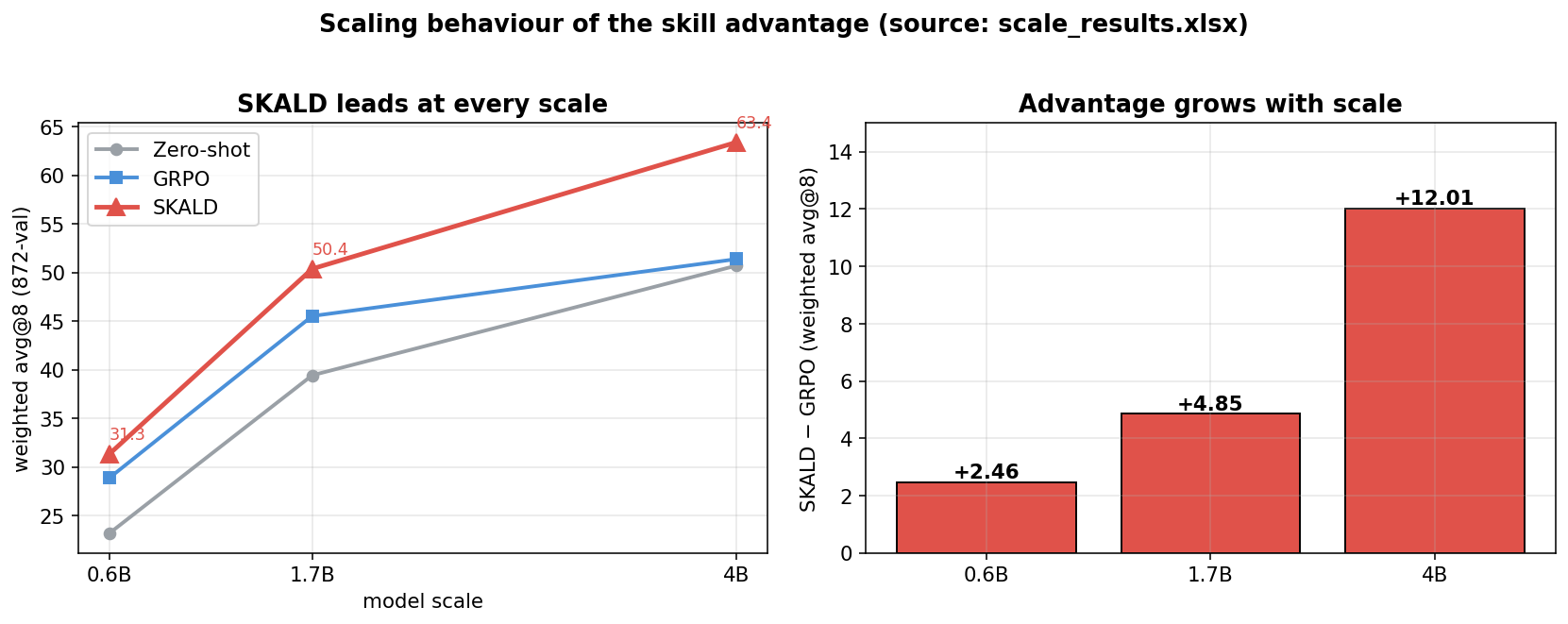}
\caption{Observed SKALD--GRPO difference in overall avg@8 at the three tested Qwen3-Base sizes. The monotonic pattern is descriptive and is not presented as a scaling law.}
\label{fig:scale}
\end{figure}

SKALD attains the best overall avg@8 and pass@8 at all three tested scales,
with mean gains over GRPO of $+2.46$, $+4.85$, and $+12.01$
(Table~\ref{tab:main}, Figure~\ref{fig:scale}). At 1.7B, the per-seed
differences are $+4.70$, $+4.91$, and $+4.94$; the hierarchical 95\% interval
is $[+3.92,+5.78]$. Rollout-matched GRPO uses $88.0$ GPU-hours and reaches
$45.52$; an equal-FLOP/equal-GPU-hour run uses $131.2$ GPU-hours and reaches
$46.31$, versus SKALD's $50.37$. At 4B, tuned equal-FLOP GRPO reaches
$52.19\pm0.18$, while SKALD reaches $63.38\pm0.32$
(Supplementary Sections~2.1, 2.6, and~2.8).

The central mechanism is directly visible in the rollout groups.
Zero-variance groups account for $68.0\%$, $63.0\%$, and $63.9\%$ at 0.6B,
1.7B, and 4B; with scale, their composition shifts from all-wrong
($54.2\%\to21.8\%$) toward all-correct ($13.8\%\to42.1\%$). At 1.7B,
zero-variance-only distillation reaches $49.63$ ($+4.11$ over GRPO),
recovering $84.7\%$ of the full gain, while mixed-only distillation reaches
$46.31$ ($+0.79$). This supports the claim that SKALD primarily adds signal
where group-relative advantages vanish (Supplementary Section~2.7).

Against privileged baselines at 1.7B, SKALD is $+1.86$ above OPSD and
$+3.04$ above EOPD. The full 4B ablation is also monotonic under one addition
order: OPSD $55.67$, skill abstraction $57.32$, tilted objective $61.05$,
and the gate $63.38$. These are ordered marginal effects rather than
interaction-free contributions.
\subsection{Robustness to Alternative Explanations}
\paragraph{Leakage and memorization.}
The conditional audit shows a modest $+2.7$-point recovery increase from
question-only to question+matched-skill, far below the $+91.8$ increase from
a full solution. More importantly, deleting instance-specific numerals,
entities, and intermediate expressions preserves the downstream score
($50.24$ versus $50.37$), and replacing each card by a peer card from the
same abstract-method cluster changes overall avg@8 by only
$0.11\pm0.19$. The median skill is reused by five problems, while
problem-, solution-, and skill-level decontamination removes all flagged
cross-split overlaps before retraining. These controls do not prove zero
information leakage, but jointly make direct answer transfer an implausible
explanation for a $+4.85$-point gain.

\paragraph{Compute and baseline quality.}
The equal-compute comparison gives GRPO the entire measured SKALD budget,
including the additional scoring and gate costs. Its gain from the extra
$181$ updates is $+0.79$, leaving SKALD $+4.06$ ahead at identical FLOPs and
GPU-hours. At 4B, a scale-specific GRPO sweep raises the equal-FLOP baseline
to $52.19$, but the remaining SKALD margin is still $+11.19$. Full-solution
OPSD also consumes slightly more measured compute than SKALD at 1.7B and
reaches $48.51$, so the result is not explained by privileged scoring alone.

\paragraph{Run uncertainty and gate choice.}
All three paired 1.7B seeds yield gains between $+4.70$ and $+4.94$; the
seed$\times$problem interval excludes zero without treating the $872$
problems as independent training runs. Varying $K$ from $4$ to $32$, using
a soft gate, or refreshing it periodically/with an EMA changes avg@8 by at
most $0.39$. Thus finite-$K$ noise and teacher drift are measurable---notably,
$14.6\%$ of default positive gates reverse sign---but do not account for the
headline improvement in the tested configurations.

\begin{table}[t]
\centering
\small
\setlength{\tabcolsep}{4pt}
\begin{tabular}{lccc}
\toprule
Configuration & Skill & Tilt & avg@8 \\
\midrule
OPSD ($s{=}y^{*}$, $\tau{=}0$)
& \ding{55} & \ding{55}
& 48.51\std{0.28} \\
\quad + skill abstraction ($s$: card)
& \ding{51} & \ding{55}
& 49.06\std{0.30} \\
\quad + tilted risk ($\tau{:}0.8{\to}0$)
& \ding{51} & \ding{51}
& 49.94\std{0.33} \\
\quad + usefulness gate (= SKALD)
& \ding{51} & \ding{51}
& \textbf{50.37}\std{0.34} \\
\bottomrule
\end{tabular}
\caption{Ordered component ablation at 1.7B (three seeds, eight decodings, $T=0.7$, top-$p=0.95$). Mean $\pm$ cross-seed standard deviation. Sequential changes are $+0.55$, $+0.88$, and $+0.43$; they need not be additive under another order. Seed-level statistics are in Supplementary Section~2.1.}
\label{tab:ablation}
\end{table}
\begin{figure}[!t]
\centering
\includegraphics[width=0.45\textwidth]{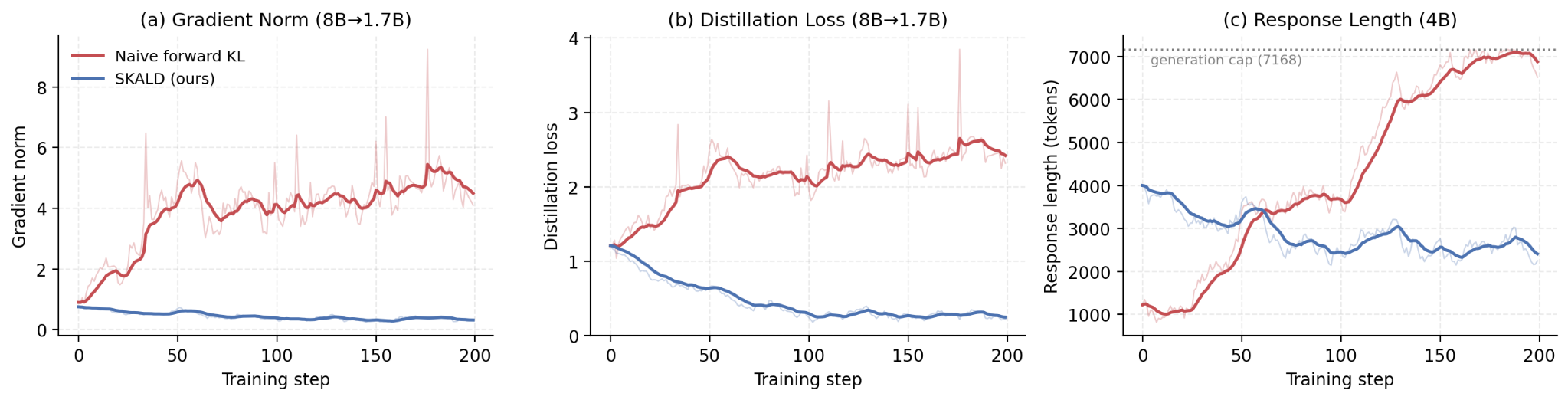}
\caption{Training dynamics in the reported risk-neutral and annealed-tilt runs. (a) Gradient norm and (b) distillation cross-entropy at 1.7B. (c) At 4B, the risk-neutral run reaches the 7168-token cap, whereas the tilted run remains shorter. These trajectories are consistent with a stabilization effect but do not replace replicated causal controls.}
\label{fig:dynamics}
\end{figure}
\begin{figure}[!t]
\centering
\includegraphics[width=0.45\textwidth]{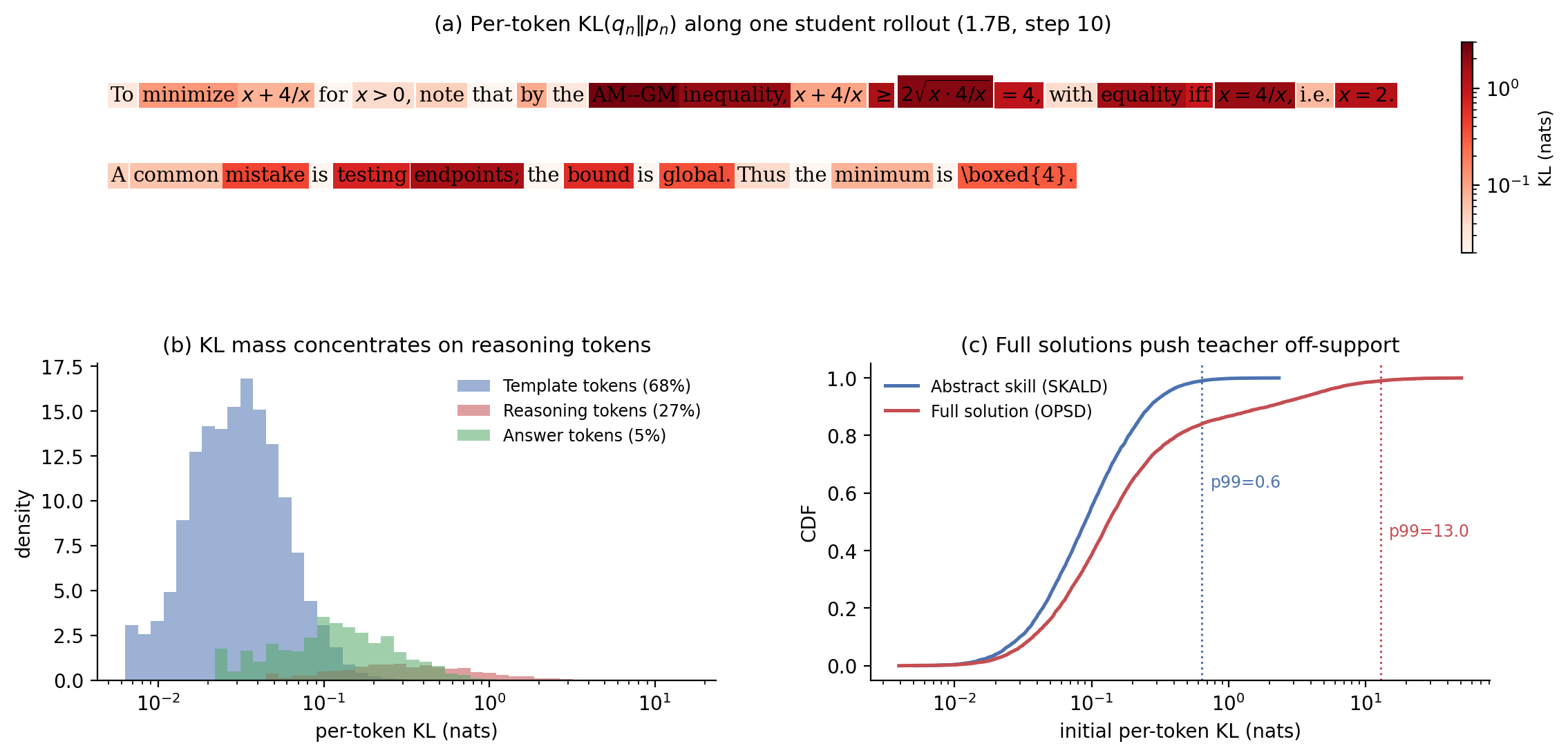}
\caption{Diagnostics for nontriviality and teacher--student mismatch. (a) One illustrative rollout, which is not used as population-level evidence. (b) Aggregate KL by token category under the stated annotation protocol. (c) Initial per-token KL CDF in the analyzed sample: p99 is $13.0$ nats with full-solution conditioning and $0.6$ with skill conditioning.}
\label{fig:h1}
\end{figure}
\subsection{Dissecting the Gain}
\paragraph{Ordered component ablation.}
Table~\ref{tab:ablation} reports marginal changes under one addition order: skill abstraction $+0.55$, annealed tilting $+0.88$, and the gate $+0.43$. The same ordering remains positive at 4B ($+1.65$, $+3.73$, and $+2.33$). Conditional audits and sanitized cards address the main leakage alternative, while the single addition order still permits component interactions.
\paragraph{Internalization beats context.}
Should a skill's value live in the inference-time context or in the weights? We compare three 1.7B conditions under an identical no-skill-at-test protocol (Table~\ref{tab:main}): GRPO ($45.52$), GRPO with the skill in the RL context and withdrawn at test ($46.60$), and SKALD ($50.37$). This comparison contrasts contextual skill exposure with
teacher-only skill distillation under the same no-skill-at-test
evaluation, with SKALD improving by $+3.77$. The context condition remains above GRPO. The supported conclusion is therefore limited to this protocol: teacher-only skill distillation outperforms student-context exposure with later withdrawal. Claims about complete assimilation would require direct parameter- or behavior-level probes.
\paragraph{The skill must induce a useful teacher shift.}
Removing the skill from the teacher context while keeping
the same Qwen3-Base checkpoint and shared-parameter training
setup reduces the 1.7B result from 50.37 to 48.01.
Because the teacher and student use identical parameters,
this comparison isolates the effect of privileged skill
conditioning rather than teacher capacity or model
initialization. With the skill, the context-induced
distribution shift provides a nontrivial distillation signal;
without it, the teacher and student distributions become
substantially closer.
\subsection{Stability Diagnostics}
Figure~\ref{fig:dynamics} compares the risk-neutral and tilted runs; full schedule, gradient, length, and seed-wise statistics are in Supplementary Section~2.5. At 1.7B, the risk-neutral run reaches gradient norms near
${\sim}4.5$ with heavy-tailed spikes and increasing distillation
cross-entropy, whereas the tilted run remains near $0.3$ and its monitored
cross-entropy falls from $1.21$ to $0.23$. This pattern is consistent with
Lemma~\ref{lem:tilt}: $r_\tau$ initially removes mass from low-$p$ teacher
tokens and gradually restores it as $\tau$ decays.

At 4B, the risk-neutral run grows from $1.2$k to $6.8$k tokens, hits the
$7168$ cap, and scores $48.18$; the tilted run contracts to roughly
$2.4$k and scores $63.38$. The $1{,}000$-update statistics come from
separate long-horizon 1.7B diagnostic runs; all main experiments use
370 updates. Their gradient-norm p95/max fall from $6.42/27.4$ to
$0.61/1.42$. Risk-neutral training diverges in $2/3$ 1.7B and $3/3$ 4B
seeds, whereas all tilted runs converge. Cosine and exponential
schedules differ from linear by at most $0.08$, and scores vary by at
most $0.15$ over $\tau_0\in[0.6,1.0]$; constant $\tau=0.8$ and
risk-neutral $\tau=0$ lose $0.62$ and $5.47$ points. These diagnostics
support stabilization; guarantees apply only to logit gradients, while
parameter-gradient bounds require a bounded Jacobian.

\subsection{Diagnostic Evidence}
Figure~\ref{fig:h1} examines where the two context views differ and
how large the initial mismatch is.
\paragraph{Token location.}
Panel (a) is an illustrative rollout: KL is small on template tokens and
peaks near the AM--GM decision, equality condition, and a skill-flagged
mistake. It should not be used alone to claim a general concentration
pattern. In the aggregate token set summarized by panel (b), the stated
reasoning category covers $27\%$ of positions and carries substantially
more KL mass than the template category. A reproducible claim requires the
number of problems, tokens, seeds, category definition, and annotation
agreement.
\paragraph{Context mismatch.}
In the analyzed initial-token sample, skill conditioning has p99 KL of
$0.6$ nats, while full-solution conditioning has p99 KL of $13.0$ nats.
This supports the narrower claim that the sampled full-solution context
induces a heavier mismatch tail. It does not establish that every such
token is unreachable, that the tilted objective transfers all skill tokens,
or that this diagnostic alone causes the downstream score difference.

\section{Conclusion}
SKALD transfers a reusable skill abstraction from a privileged context view
into a question-only policy, without a larger online teacher or privileged
input at inference. Its annealed R\'enyi-type cross-entropy makes the
optimization trade-off explicit: positive tilt discounts low-student-mass
teacher tokens and targets a sharpened escort distribution; annealing to
zero restores teacher cross-entropy and the forward-KL student gradient.

The new controls connect this objective to the motivating failure mode.
Zero-variance groups form $63$--$68\%$ of training groups, and restricting
distillation to them recovers $84.7\%$ of the full 1.7B gain. SKALD remains
$+4.06$ above equal-FLOP GRPO at 1.7B and $+11.19$ above tuned equal-FLOP
GRPO at 4B; the latter scale also preserves positive skill, tilt, and gate
increments. Question-conditioned recovery, card sanitization, human audit,
and problem/solution/skill decontamination make direct answer leakage an
unlikely explanation, while hierarchical intervals preserve the gain across
seeds and evaluation problems. Across the three tested Qwen3-Base sizes,
these results support skill-conditioned self-distillation as a compute-aware
complement to RLVR precisely where group-relative rewards are silent.

\bibliography{aaai2027}


\end{document}